%% file: main.tex
\documentclass{article}

\usepackage[accepted]{icml2025}
\usepackage{times}

\input{headers/math_commands.tex}

\input{headers/header-lqa}

\input{headers/header}

\input{headers/header-lqa-tail}

\icmltitlerunning{Proof or Bluff? Evaluating LLMs on 2025 USA Math Olympiad}

\begin{document}

\twocolumn[
\icmltitle{Proof or Bluff? Evaluating LLMs on 2025 USA Math Olympiad}

\icmlsetsymbol{equal}{*}

\begin{icmlauthorlist}
\icmlauthor{Ivo Petrov}{insait}
\icmlauthor{Jasper Dekoninck}{eth}
\icmlauthor{Lyuben Baltadzhiev}{insait}
\icmlauthor{Maria Drencheva}{insait}
\icmlauthor{Kristian Minchev}{insait}
\icmlauthor{Mislav Balunović}{insait,eth}
\icmlauthor{Nikola Jovanović}{eth}
\icmlauthor{Martin Vechev}{insait,eth}
\end{icmlauthorlist}

\icmlaffiliation{insait}{INSAIT, Sofia University "St. Kliment Ohridski"}
\icmlaffiliation{eth}{ETH Zurich}

\icmlcorrespondingauthor{Ivo Petrov}{ivo.petrov@insait.ai}

\icmlkeywords{Machine Learning, Artificial Intelligence, AI, Reasoning, Proofs, Maths, Competitions, LLM, ICML}
\vskip 0.3in
]
\printAffiliationsAndNotice{}

\input{paper_files/abstract}

\input{paper_files/introduction}

\input{paper_files/related.tex}

\input{paper_files/methodology.tex}

\input{paper_files/results}

\input{paper_files/discussion.tex}

\input{paper_files/limitations.tex}

\input{paper_files/conclusion}

\input{paper_files/acknowledgements}

\message{^^JLASTBODYPAGE \thepage^^J}

\bibliography{paper_files/references}
\bibliographystyle{plainnat}

\message{^^JLASTREFERENCESPAGE \thepage^^J}

\ifbool{includeappendix}{%
	\clearpage
	\onecolumn
	\appendix
	\input{paper_files/appendix}

}{}

\message{^^JLASTPAGE \thepage^^J}

\end{document}

%% file: headers/math_commands.tex
\usepackage{amsmath,amsfonts,bm}

\def\eqref#1{equation~\ref{#1}}

\def\1{\bm{1}}

\DeclareMathAlphabet{\mathsfit}{\encodingdefault}{\sfdefault}{m}{sl}
\SetMathAlphabet{\mathsfit}{bold}{\encodingdefault}{\sfdefault}{bx}{n}

%% file: headers/header-lqa.tex
\usepackage[utf8]{inputenc} %
\usepackage{lipsum} %

\usepackage[T1]{fontenc}

\usepackage{etoolbox}
\newbool{includeappendix}
\setbool{includeappendix}{true}

\input{headers/lqa/overfull}

\input{headers/lqa/comments}
\input{headers/lqa/colors}

\input{headers/lqa/listing}

%% file: headers/lqa/overfull.tex
\ifdefined\isoverfull
\else
\fi

%% file: headers/lqa/colors.tex
\usepackage{xcolor} %

\definecolor{my-full-blue}{HTML}{1F77B4}

\definecolor{my-full-orange}{HTML}{FF7F0E}

\definecolor{my-full-green}{HTML}{2CA02C}

\definecolor{my-full-red}{HTML}{d62728}

\definecolor{my-full-purple}{HTML}{9467bd}

\colorlet{my-blue}{my-full-blue!30}
\colorlet{my-orange}{my-full-orange!30}
\colorlet{my-green}{my-full-green!30}
\colorlet{my-red}{my-full-red!30}
\colorlet{my-purple}{my-full-purple!30}

%% file: headers/lqa/listing.tex
\usepackage{listings}

\usepackage{textcomp}

\usepackage{xcolor}

\usepackage[scaled=0.8]{beramono}

\definecolor{ckeyword}{HTML}{7F0055}
\definecolor{ccomment}{HTML}{3F7F5F}
\definecolor{cstring}{HTML}{2A0099}

\lstdefinestyle{numbers}{
	numbers=left,
	framexleftmargin=20pt,
	numberstyle=\tiny,
	firstnumber=auto,
	numbersep=1em,
	xleftmargin=2em
}

\lstdefinestyle{layout}{
	frame=none,
	captionpos=b,
}

\lstdefinestyle{comment-style}{
	morecomment=[l]//,
	morecomment=[s]{/*}{*/},
	commentstyle={\color{ccomment}\itshape},
}

\lstdefinestyle{string-style}{
	morestring=[b]",%
	morestring=[b]',%
	stringstyle={\color{cstring}},
	showstringspaces=false,%
}

\lstdefinestyle{keyword-style}{
	keywordstyle={\ttfamily\bfseries},
	morekeywords={
		function,
		constructor,
		int,
		bool,
		return,
		returns,
		uint
	},
	morekeywords = [2]{},
	keywordstyle = [2]{\text},
	sensitive=true,
}

\lstdefinestyle{input-encoding}{
	inputencoding=utf8,
	extendedchars=true,
	literate=
	{ℝ}{$\reals$}1%
	{→}{$\rightarrow$}1%
	{α}{$\alpha$}1%
	{β}{$\beta$}1%
	{λ}{$\lambda$}1%
	{θ}{$\theta$}1%
	{ϕ}{$\phi$}1%
}

\lstdefinestyle{escaping}{
	moredelim={**[is][\color{blue}]{\%}{\%}},
	escapechar=|,
	mathescape=true
}

\lstdefinestyle{default-style}{
	basicstyle=\fontencoding{T1}\ttfamily\footnotesize,
	style=numbers,
	style=layout,
	style=comment-style,
	style=string-style,
	style=keyword-style,
	style=input-encoding,
	style=escaping,
	tabsize=2,
	upquote=true
}

\lstdefinelanguage{BASIC}{
	language=C++,
	style=default-style
}[keywords,comments,strings]%

%% file: headers/header.tex
\usepackage{booktabs}       %
\usepackage{nicefrac}       %
\usepackage[flushleft]{threeparttable}

\usepackage{url}
\usepackage{xcolor}
\usepackage{xspace}
\usepackage{etoolbox}
\usepackage{fontawesome}
\usepackage{enumitem}
\usepackage{amsmath,amssymb}
\usepackage{array}
\usepackage{multirow}
\usepackage{wrapfig}
\usepackage{adjustbox,booktabs}
\usepackage{siunitx}
\usepackage{caption}
\usepackage{fontawesome}
\usepackage{stackengine}
\usepackage{svg}
\usepackage{mathtools}
\usepackage{xspace}
\usepackage{wrapfig}
\usepackage{makecell}
\usepackage{graphicx}
\usepackage{booktabs}
\usepackage{longtable}
\usepackage{hyperref}
\usepackage[labelformat=simple]{subcaption}
\usepackage[most]{tcolorbox}

\usepackage{colortbl}

\input{headers/lqa/tikz}

\usepackage{amsthm}
\usepackage{stmaryrd}

\newcolumntype{x}[2]{S[table-format=#1.#2,table-auto-round]}
\newcolumntype{M}{>{$}l<{$}}

\NewDocumentCommand{\phimodel}{O{}}{%
\textsc{Phi}\ifstrempty{#1}{}{-{#1}}\xspace
}

\newcommand{\ofour}{\textsc{o4-mini}\xspace}
\newcommand{\othree}{\textsc{o3}\xspace}
\newcommand{\othreemini}{\textsc{o3-mini}\xspace}
\newcommand{\oone}{\textsc{o1-pro}\xspace}
\newcommand{\rone}{\textsc{DeepSeek-R1}\xspace}

\newcommand{\qwq}{\textsc{QwQ}\xspace}
\newcommand{\grok}{\textsc{Grok 3}\xspace}

\newcommand{\flthink}{\textsc{Flash-Thinking}\xspace}
\newcommand{\claude}{\textsc{Claude 3.7}\xspace}
\newcommand{\geminipro}{\textsc{Gemini-2.5-Pro}\xspace}

\definecolor{acceptblue}{HTML}{6494EA}
\hypersetup{citecolor=acceptblue}

\definecolor{lightred}{HTML}{ffcbc7}
\definecolor{gemini}{HTML}{4285F4}
\definecolor{claude}{HTML}{f3e9d7}
\definecolor{deepseek}{HTML}{FADA4B}
\definecolor{qwen}{HTML}{FA574B}
\definecolor{oai}{HTML}{10a37f}
\definecolor{LightGreen}{HTML}{CCFFCC}

\lstdefinestyle{mystyle}{
    breaklines=true,
    basicstyle=\scriptsize\ttfamily,
    numbers=none,
    language={},
    framextopmargin=0pt,
    framexbottommargin=0pt,
    breakindent=0pt,
    showspaces = false,
    keywordstyle=\bfseries,
    showstringspaces=false,
    columns=fullflexible,
    morekeywords={Answer}
    moredelim=[**][\bfseries]{!!}
}

\newtcblisting{prompt}[2][]{
    arc=3pt, outer arc=3pt,
    width=\linewidth,
    left=1mm,
    top=0mm,
    bottom=0mm,
    title=#2, 
    colback=lightred!5!white,
    colframe=black,
    fonttitle=\bfseries,
    listing only, 
    listing options={style=mystyle},
    breakable,
    #1
}

\newtcblisting{problem}[2][]{
    arc=3pt, outer arc=3pt,
    width=\linewidth,
    left=1mm,
    top=0mm,
    bottom=0mm,
    title=#2, 
    colback=acceptblue!5!white,
    colframe=acceptblue!50!black,
    fonttitle=\bfseries,
    listing only, 
    listing options={style=mystyle},
    breakable,
    #1
}

\newtcblisting{gpt}[2][]{
    arc=3pt, outer arc=3pt,
    width=\linewidth,
    left=1mm,
    top=0mm,
    bottom=0mm,
    title=#2, 
    colback=oai!5!white,
    colframe=oai!50!black,
    fonttitle=\bfseries,
    listing only, 
    listing options={style=mystyle},
    breakable,
    #1
}

\newtcblisting{claudels}[2][]{
    arc=3pt, outer arc=3pt,
    width=\linewidth,
    left=1mm,
    top=0mm,
    bottom=0mm,
    title=#2, 
    colback=claude!5!white,
    colframe=claude!85!black,
    fonttitle=\bfseries,
    listing only, 
    listing options={style=mystyle},
    breakable,
    #1
}

\newtcblisting{deepseekls}[2][]{
    arc=3pt, outer arc=3pt,
    width=\linewidth,
    left=1mm,
    top=0mm,
    bottom=0mm,
    title=#2, 
    colback=deepseek!5!white,
    colframe=deepseek!85!black,
    fonttitle=\bfseries,
    listing only, 
    listing options={style=mystyle},
    breakable,
    #1
}

\newtcblisting{geminils}[2][]{
    arc=3pt, outer arc=3pt,
    width=\linewidth,
    left=1mm,
    top=0mm,
    bottom=0mm,
    title=#2, 
    colback=gemini!5!white,
    colframe=gemini!50!black,
    fonttitle=\bfseries,
    listing only, 
    listing options={style=mystyle},
    breakable,
    #1
}

\newtcblisting{geminilsescaped}[2][]{
    arc=3pt, outer arc=3pt,
    width=\linewidth,
    left=1mm,
    top=0mm,
    bottom=0mm,
    title=#2, 
    colback=gemini!5!white,
    colframe=gemini!50!black,
    fonttitle=\bfseries,
    listing only, 
    listing options={style=mystyle, escapeinside=||},
    breakable,
    #1
}

\newtcblisting{qwenls}[2][]{
    arc=3pt, outer arc=3pt,
    width=\linewidth,
    left=1mm,
    top=0mm,
    bottom=0mm,
    title=#2, 
    colback=qwen!5!white,
    colframe=qwen!50!black,
    fonttitle=\bfseries,
    listing only, 
    listing options={style=mystyle},
    breakable,
    #1
}

\newcommand{\othreeheader}{
    \begin{tikzpicture}
      \node[anchor=north] {\pgftext{\includegraphics[width=0.34cm]{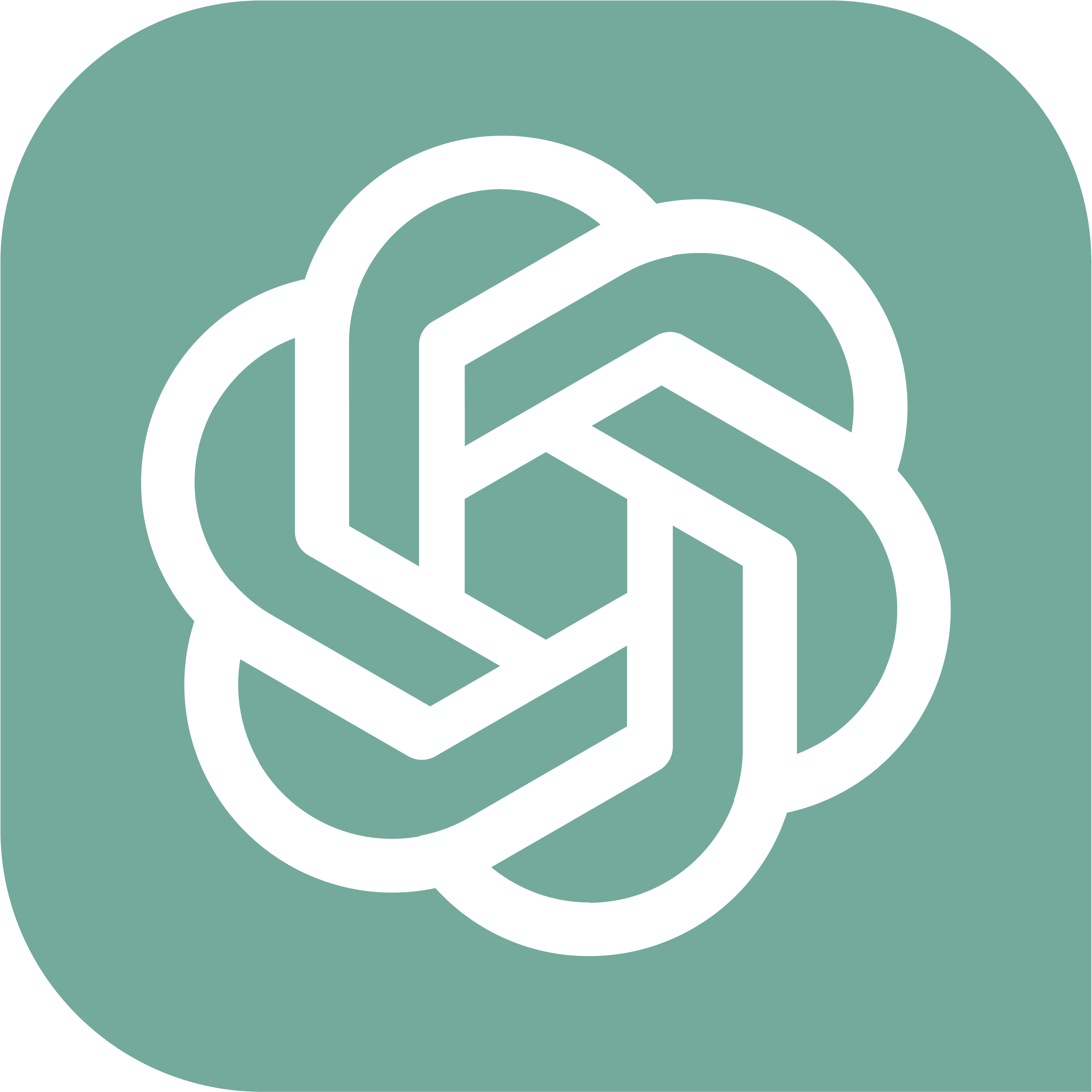}}};
    \end{tikzpicture}
    ~\othreemini{}
}

\newcommand{\geminiheader}{
    \begin{tikzpicture}
      \node[anchor=north] {\pgftext{\includegraphics[width=0.34cm]{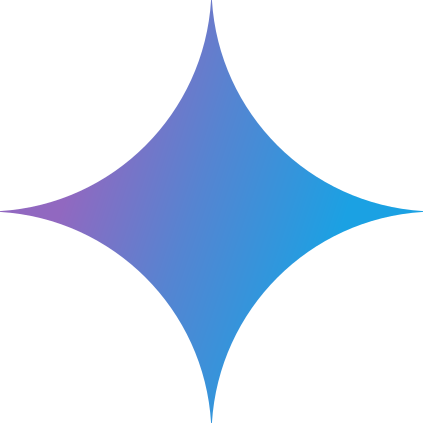}}};
    \end{tikzpicture}
    ~\flthink{}
}

\newcommand{\geminiproheader}{
    \begin{tikzpicture}
      \node[anchor=north] {\pgftext{\includegraphics[width=0.34cm]{figures/assets/Gemini_2024_icon.png}}};
    \end{tikzpicture}
    ~\geminipro{}
}

\newcommand{\claudeheader}{
    \begin{tikzpicture}
      \node[anchor=north] {\pgftext{\includegraphics[width=0.34cm]{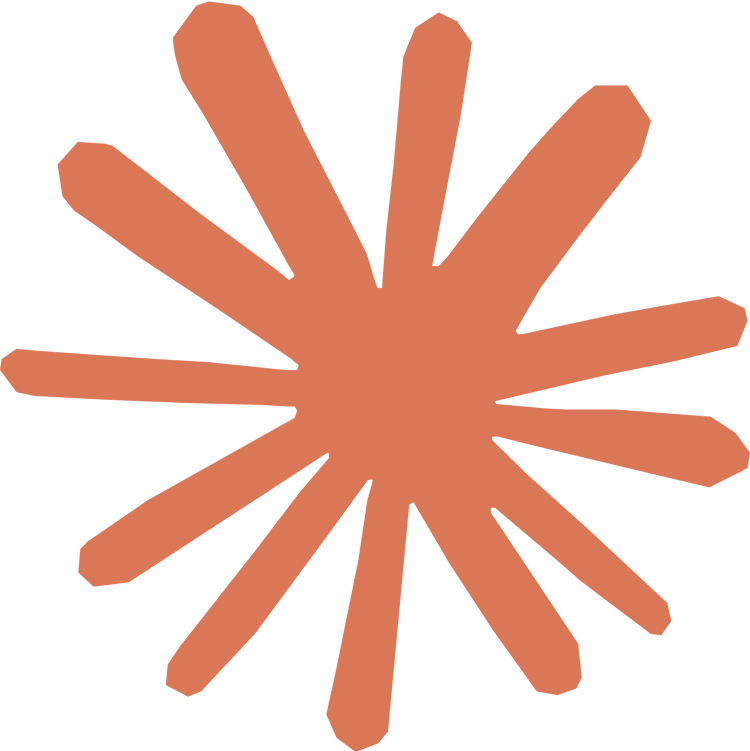}}};
    \end{tikzpicture}
    ~\claude{} 
}

\newcommand{\qwenheader}{
    \begin{tikzpicture}
      \node[anchor=north] {\pgftext{\includegraphics[width=0.34cm]{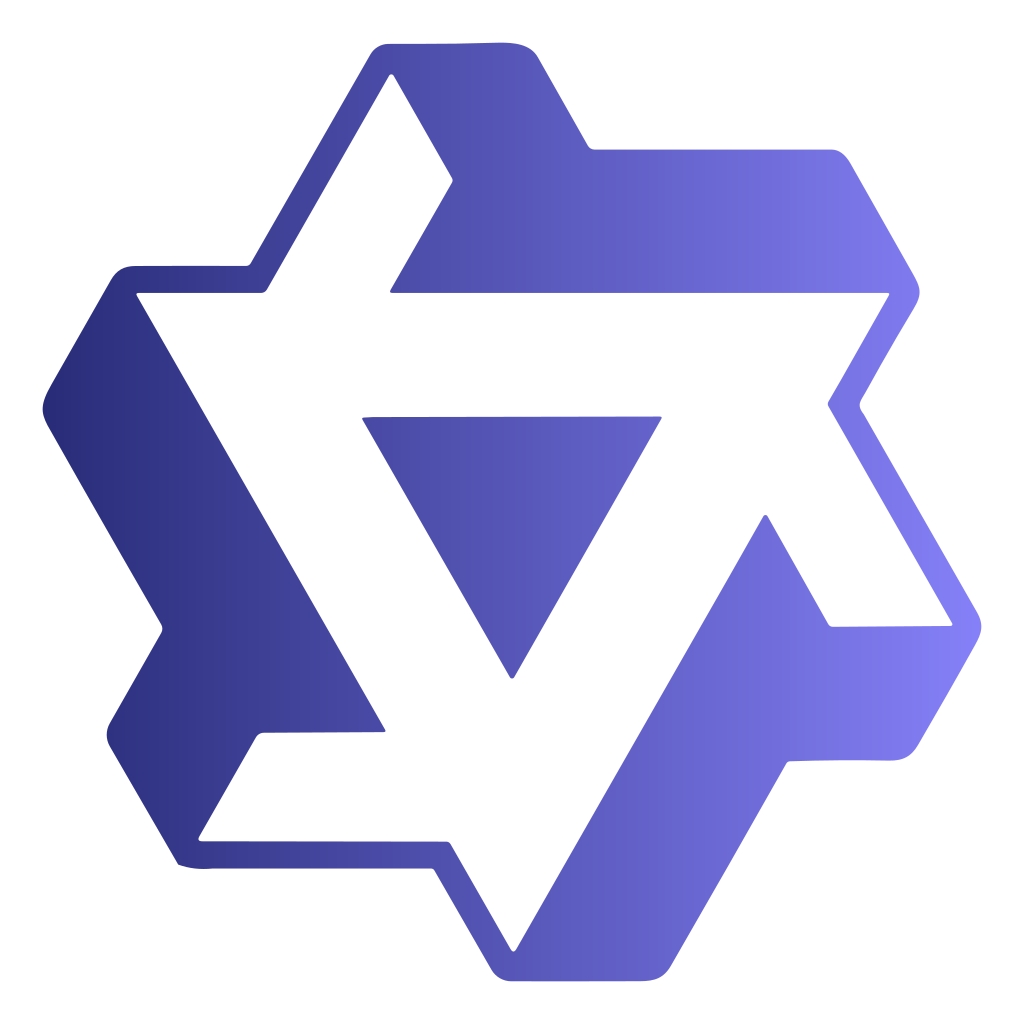}}};
    \end{tikzpicture}
    ~\qwq{} 
}

\newcommand{\deepseekheader}{
    \begin{tikzpicture}
      \node[anchor=north] {\pgftext{\includegraphics[width=0.34cm]{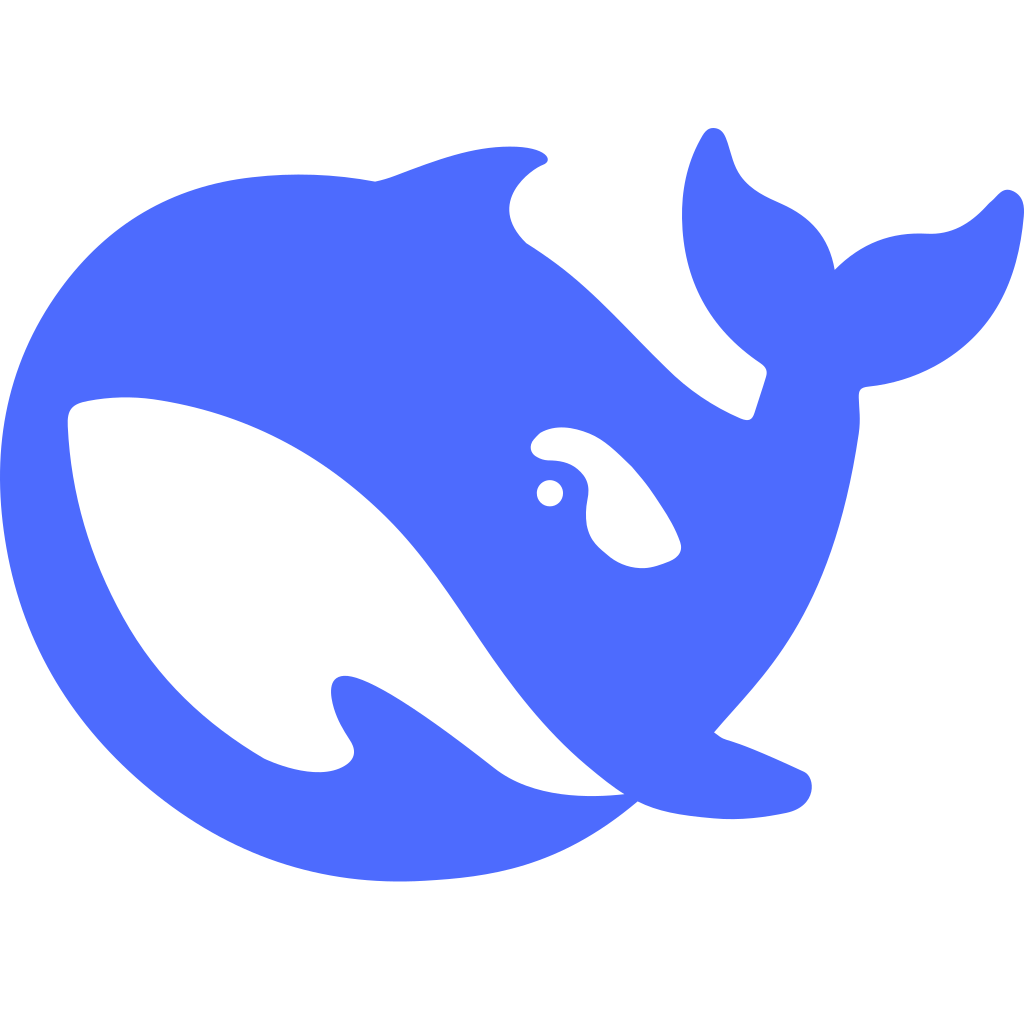}}};
    \end{tikzpicture}
    ~\rone{} 
}

%% file: headers/lqa/tikz.tex
\usepackage{tikz}

\usetikzlibrary{calc,decorations,decorations.pathmorphing}
\usetikzlibrary{positioning,fit,arrows}
\usetikzlibrary{decorations.markings}
\usetikzlibrary{shapes,shapes.geometric}
\usetikzlibrary{shadows,patterns,snakes}
\usetikzlibrary{backgrounds,decorations.pathreplacing,calligraphy,automata}
\usetikzlibrary{intersections}
\usetikzlibrary{angles,quotes}
\usetikzlibrary{plotmarks}
\usetikzlibrary{patterns}
\usetikzlibrary{arrows.meta}
\usetikzlibrary{shapes.misc}
\usetikzlibrary{chains}

%% file: headers/header-lqa-tail.tex
\input{headers/lqa/references}

%% file: headers/lqa/references.tex
\usepackage[capitalize]{cleveref}

\crefformat{section}{\S#2#1#3}

\crefrangeformat{section}{\S#3#1#4\crefrangeconjunction\S#5#2#6}

\crefmultiformat{section}{\S#2#1#3}{\crefpairconjunction\S#2#1#3}{\crefmiddleconjunction\S#2#1#3}{\creflastconjunction\S#2#1#3}

\newcommand{\crefrangeconjunction}{--}

\crefname{listing}{Lst.}{listings}
\crefname{line}{Lin.}{Lin.}
\crefname{appendix}{App.}{App.}

\newcommand{\app}[1]{%
	\ifbool{includeappendix}{\cref{#1}}{the appendix}%
}
\newcommand{\App}[1]{%
	\ifbool{includeappendix}{\cref{#1}}{The appendix}%
}

%% file: paper_files/abstract.tex
\begin{abstract}
    Recent mathematical benchmarks indicate that large language models (LLMs) achieve strong performance in mathematical competitions such as AIME, with leading models attaining scores comparable to or exceeding those of top human participants. However, these benchmarks evaluate models solely based on final numerical answers, neglecting rigorous reasoning and proof generation which are essential for real-world mathematical tasks. To address this, we introduce a comprehensive evaluation of full-solution reasoning for challenging mathematical problems. Using expert human annotators, we evaluated several state-of-the-art reasoning models on the six problems from the 2025 USAMO within hours of their release. Our results show that all tested models struggle significantly, with none exceeding a score of $30\%$, and most achieving only trivial scores below $5\%$. Through a detailed analysis of reasoning traces, we identify the most common failure modes and find several unwanted artifacts arising from the optimization strategies employed during model training. Overall, our results underscore the limitations of current LLMs in tasks requiring deep mathematical understanding and emphasize the need for significant advances in reasoning and proof generation capabilities.
\end{abstract}

%% file: paper_files/introduction.tex
\vspace{-5mm}
\section{Introduction}
The advancement of reasoning models has significantly improved the mathematical capabilities of large language models (LLMs). Evaluation efforts demonstrate that these models achieve impressive performance in mathematical competitions such as AIME \citep{deepmind2025geminipro,openai2025introducing}. However, these competitions only evaluate final numerical answers and do not require rigorous proof-based reasoning essential for most mathematical tasks. 
 
\vspace{-1mm}
\paragraph{Benchmarking mathematical reasoning} Current benchmarks that mitigate this issue either rely on formal verification tools such as Lean \citep{minif2f,fimo,putnambench} or focus on the evaluation of constructive proofs \citep{mathconstruct}. While these approaches are useful, the former does not leverage the LLMs' strong natural language generation capabilities, and the latter covers only a limited subset of proofs. Only \citet{frieder2024} and \citet{brains} evaluate LLMs on full-solution reasoning, but both of these evaluations are limited to a static analysis of mathematical capabilities. As a result, neither presents their work as a benchmark, and both include problems that could potentially have appeared in the training data. Therefore, it remains uncertain whether current reasoning LLMs can reliably address complex mathematical questions requiring rigorous reasoning, which is crucial in real-world mathematical contexts.

\vspace{-2mm}
\paragraph{Our approach} To overcome these limitations, we conduct the first evaluation of natural language proofs by LLMs on challenging problems from the 2025 USA Mathematical Olympiad (USAMO). The USAMO represents one of the highest tiers of high school mathematics competitions in the United States, demanding detailed proofs and explanations analogous to the International Mathematical Olympiad (IMO). Participants qualify through prior competitions, including the AIME, but USAMO problems require significantly more rigorous solutions.

Overall, we find that current LLMs struggle significantly on USAMO problems, with the best-performing model achieving an average score of around $30\%$, and the majority obtaining less than $5\%$. Our evaluation reveals several critical failure modes, including flawed logic, unjustified assumptions, and a lack of creativity. These findings underscore the substantial limitations of current LLMs in generating rigorous proofs. In this report, we first describe the landscape of mathematical benchmarking in \cref{sec:related}, outline our evaluation methodology in \cref{sec:meth}, present detailed results and identify critical weaknesses in \cref{sec:results}, and discuss qualitative observations in \cref{sec:discussion}.

\vspace{-2mm}
\paragraph{Main contributions} Our key contributions include:
\vspace{-3mm}
\begin{itemize}\setlength\itemsep{0.005em}
    \item A standardized framework for unbiased evaluation of mathematical proofs.
    \item A thorough evaluation of 11 of the best reasoning LLMs on the USAMO 2025 competition.
    \item A qualitative analysis of common artifacts and issues in the solution traces.
\end{itemize}

%% file: paper_files/related.tex
\section{Related Work}
\label{sec:related}
The benchmarking of mathematical reasoning in LLMs can be roughly divided into three categories: answer-based benchmarks, formal verification, and full-proof evaluation. We discuss each of these approaches below.
\vspace{-1mm}

\paragraph{Answer-based Benchmarks}
The most prominent benchmarks for mathematical reasoning evaluate problems by comparing numerical or algebraic expressions to a ground truth. Early datasets like GSM8K \citep{gsm8k} and MATH \citep{math500} are effectively “solved” by the latest large-language models \citep{r1,qwq32b}, and even more advanced collections of competition-style problems such as Omni-MATH \citep{omnimath}, OlympiadBench \citep{olympiadbench}, and AIME 2024/25, are now approaching saturation with the release of systems such as \geminipro{} and \ofour{}. Moreover, these benchmarks fail to penalize flawed reasoning, as models can produce the correct final answer despite serious logical errors \citep{rstar}. Since all of these datasets are publicly accessible, there is also a significant risk of inadvertent data contamination.
\vspace{-1mm}

Private benchmarks such as FrontierMath \citep{frontiermath} mitigate leakage and present substantially tougher problems, but their hidden nature hinders transparent tracking of progress in the field. Furthermore, the benchmark also relies on answer-based evaluation, which does not capture the full complexity of mathematical reasoning.

\vspace{-1mm}

\paragraph{Formal Verification}
An alternative evaluation approach relies on formal proof verification. These benchmarks require models to generate proofs in formal languages such as Lean or Isabelle, where correctness can be rigorously validated~\citep{putnambench, minif2f}. However, while this approach ensures rigor, it struggles to leverage the natural language capabilities of LLMs, even with specialized training. Additionally, minor errors can cause an otherwise valid proof to be rejected, making this evaluation method overly punitive for mostly correct reasoning.
\vspace{-1mm}

\paragraph{Full-proof evaluation}
Studies evaluating LLMs' ability to produce complete and correct mathematical proofs are still rare. \citet{frieder2024} evaluates full-proof solutions but only on two older, non-reasoning models, providing a limited perspective on current capabilities. 

A concurrent study has also found that LLMs struggle with solving proof-based problems \citep{brains}. Their evaluation focuses on problems from the IMO Shortlist, which are comparable in difficulty to those from the USAMO. However, our work evaluates more recent LLMs that exhibit significant improvements in mathematical reasoning, such as \geminipro{} and \ofour{}. Additionally, instead of labelling each solution as simply correct or incorrect, we evaluate model responses on a continuous grading scale. This provides a more nuanced measure of performance, which is crucial in real-world use cases.

%% file: paper_files/methodology.tex
\section{Methodology}\label{sec:meth}
\vspace{-1mm}
This section outlines the methodology used for evaluating the mathematical reasoning capabilities of the selected LLMs. Our approach prioritized accuracy, fairness, and transparency throughout each step, involving careful preparation and grading procedures.

\begin{figure*}[t]
    \centering
    \includegraphics[width=0.75\linewidth]{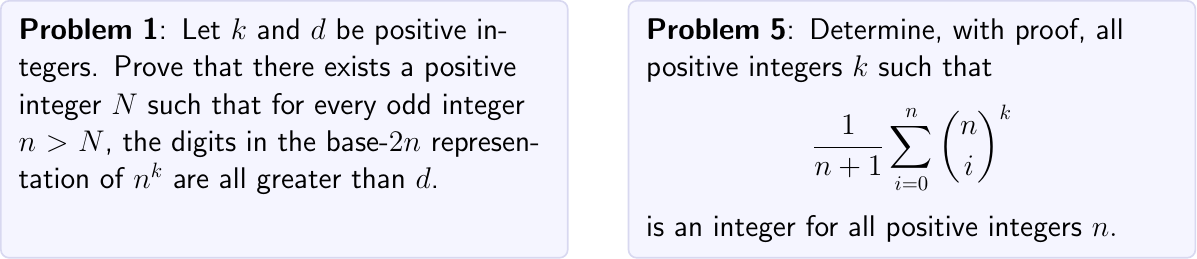}
    \vspace{-2mm}
    \caption{Two problems of USAMO 2025. The other problems are available in \cref{app:problems}}
    \label{fig:problems}
    \vspace{-0.2in}
\end{figure*}
\vspace{-1mm}
\subsection{Problem Selection and Preparation}
We selected the USAMO 2025, a highly prestigious mathematics competition comprising six proof-based problems administered over two days, as our benchmark. This competition aligns perfectly with our evaluation objectives, as the questions are challenging, require detailed proofs for full credit, and are uncontaminated. In \cref{fig:problems}, we present two problems from the competition, with the remaining four available in \cref{app:problems}.
\vspace{-1mm}

For evaluation, we provided each model with the problems, prompting them explicitly to produce comprehensive and detailed proofs formatted in \LaTeX. The full prompt instructions and details of used hyperparameters are available in \cref{app:exp_prompt}. To reduce variance, each model solved every problem four separate times. Solutions, excluding thought traces, were anonymized and converted into PDF format for grading. \grok{}, \geminipro{}, \othree{}, \ofour{}, and \rone{}~\textsc{(05/28)} were evaluated after our initial grading and were therefore not fully anonymous when presented to the judges.
\vspace{-1mm}

\subsection{Judge Selection and Training}
Our grading team consisted of four experts, each having substantial mathematical problem-solving experience as a former national IMO team member or having participated in final-stage team selection process for their respective country. Prior to the grading, judges received instructions detailing evaluation goals and methodologies. These guidelines are accessible in the supplementary material. A trial run with three USAMO 2024 problems was conducted to familiarize evaluators with the grading approach and resolve ambiguities. Small misunderstandings were clarified during this session.
\vspace{-1mm}

\subsection{Grading Procedure}

Each of the six problems from USAMO 2025 was independently evaluated by two evaluators, with each judge responsible for grading three unique problems. This double grading method, modelled after the IMO's evaluation process, ensures consistency in our grading and decreases biases.
\vspace{-1mm}

Since the official USAMO does not release standard solutions or grading schemes, we carefully developed a standardized grading scheme for each problem, drawing from reliable mathematical community resources, particularly the Art of Problem Solving (AoPS) forums. All solutions from these sources were verified by our evaluators for accuracy before creating the grading scheme. Following USAMO conventions, each solution was graded out of a maximum of seven points with partial credit given for significant and meaningful progress. The finalized grading schemes are available in the supplementary material. An example can be found in \cref{app:exp_grading}.

Judges independently reviewed each assigned solution against the pre-established grading scheme. When a solution did not perfectly align with the scheme, the approach was awarded points where appropriate. Each judge documented their reasoning, including justification for each partial credit awarded. These notes are also accessible on our website, with an example provided in \cref{app:exp_grading}.

Evaluators also documented prominent failure modes observed during grading, defined as the first instance of incorrect or inadequately explained reasoning. Specifically, mistakes were categorized into four classes:
\vspace{-1mm}
\begin{itemize}[leftmargin=25pt]
\setlength\itemsep{0.01em}
\item \textbf{Logic:} Errors due to logical fallacies or unjustified leaps disrupting the reasoning.
\item \textbf{Assumption:} Errors coming from the introduction of unproven or incorrect assumptions that undermined subsequent steps.
\item \textbf{Creativity:} Errors resulting from fundamentally incorrect solution strategies due to the inability to identify the correct approach.
\item \textbf{Algebra/Arithmetic:} Errors arising from critical algebraic or arithmetic miscalculations.
\end{itemize}
\vspace{1mm}
We show examples of these errors in \cref{app:error_mode_examples}.
\vspace{1mm}

Additionally, noteworthy behaviors or trends in model-generated solutions were systematically logged for further analysis. These observations were used to identify common pitfalls and areas for improvement in the models' reasoning capabilities and are presented in \cref{sec:discussion}.

%% file: paper_files/results.tex
\vspace{-1mm}
\section{Results}\label{sec:results}
\vspace{-1mm}
\input{tables/main_results}

We now present the results of our evaluation. In \cref{subsec:main_results}, we detail our primary findings, demonstrating that few models achieve a score above $5\%$, with \geminipro{} being the only one available before the public release of the USAMO 2025 problems that scores a non-trivial result of around $25\%$. In \cref{subsec:failure_modes}, we analyze common failure patterns in depth, identifying typical mistakes and trends in the models' reasoning. Finally, in \cref{subsec:automated}, we try and fail to automatically grade the models' solutions by giving a judge model the grading scheme and a ground-truth solution.
\vspace{-1mm}
\subsection{Main Results}\label{subsec:main_results}
\vspace{-1mm}
\paragraph{Model performance}
We evaluate eleven state-of-the-art reasoning models on the 2025 USAMO problems. Specifically, we chose \qwq{} \citep{qwq32b}, the original and May versions of \rone{} \citep{r1}, \geminipro{} \citep{deepmind2025geminipro}, \flthink{} \citep{gemini-1.5}, \oone{} \citep{o1}, \othreemini{} \citep{o3}, \othree{}, \ofour{} \citep{o4}, \grok{} \citep{xai2025grok3} and \claude{} \citep{anthropic2024claude}. For brevity, we use a shorthand notation for each model in the main text, and we refer to \cref{app:exp_abb} for the full model names. The chosen hyperparameters and prompt can be found in \cref{app:exp_prompt}.

We provide a detailed, per-problem breakdown of model performance in \cref{tab:main_results}, with average scores computed across four evaluation runs. Each USAMO problem is scored out of 7 points, with a total maximum score of 42 points per run. The table also includes the total cost of running each model over all problems and evaluation runs. If the model is free, we indicate this with "N/A".

While state-of-the-art LLMs perform comparably to top human competitors in competitions focused on numerical answers, our evaluation reveals a substantial gap in their ability to generate rigorous mathematical proofs. Only four models — \geminipro{}, \othree{}, \ofour{}, and \rone{}~\textsc{(05/28)} — scored above 5\%, with most of their points coming from Problems 1 and 4, the easiest problems. 
All other models scored below 5\%, highlighting a severe limitation in handling the complexity and rigor of USAMO-level problems. Notably, among the $168$ solutions from the lower-scoring models, the only perfect $7/7$ score was a single \grok{} attempt on Problem 1.

Although the USAMO is more challenging than previously tested competitions, the near-total failure of most models to solve more than one problem demonstrates that current LLMs are still far from capable of rigorous olympiad-level reasoning. Even the top four models, despite outperforming others, struggled significantly. This limitation suggests that existing optimization methods like GRPO \citep{grpo} may currently be insufficient for tasks requiring detailed logical precision.
\paragraph{Rank variance}
To estimate the statistical significance of our ranking, we also compute a confidence interval for the model ranks using a paired permutation test at a significance level of $\alpha=0.05$. In \cref{tab:main_results} we see that our benchmark clearly differentiates the top 4 models from the rest but is not sufficiently powerful to statistically claim anything about the ordering of the two groups. The only exception to this is that the test allows an inversion between \grok{} and \ofour{}, implying that the small sample size makes it unreliable to rank the models with high precision.
\subsection{Failure Modes}\label{subsec:failure_modes}
The most frequent failure mode among human participants is the inability to find a correct solution. Typically, human participants have a clear sense of whether they solved a problem correctly. In contrast, the evaluated LLMs almost always claimed to have solved the problems. Notable exceptions to this were 2 attempts by \othree{} on problem 2, where it conceded that it would not be able to produce a rigorous solution. This discrepancy poses a significant challenge for mathematical applications of LLMs as mathematical results derived using LLMs cannot be trusted without rigorous human validation. To further investigate this limitation, we conducted a thorough analysis of the errors identified during grading using the categories defined in \cref{sec:meth}.

\begin{figure*}[t]
    \centering
    \includegraphics[width=0.9\textwidth]{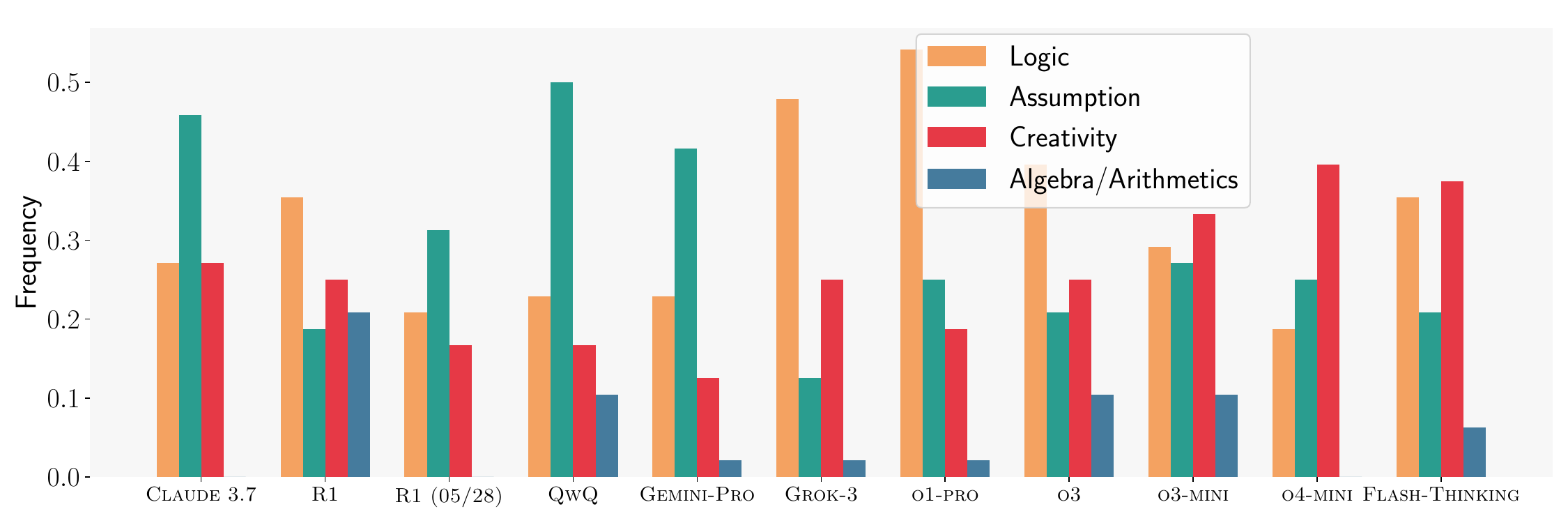}
    \vspace{-4mm}
    \caption{Distribution of first encountered failure mode.}
    \vspace{-2mm}
    \label{fig:errors}
\end{figure*}

Figure \ref{fig:errors} illustrates the distribution of these error categories as determined by our judges. The most common errors were related to flawed logic, with solutions frequently using unjustified reasoning steps, incorrect rationale, or misinterpretations of previous progress. Another significant issue was the models' tendency to treat certain critical proof steps as trivial without proper justification. Notably, \othreemini{}, despite being one of the best reasoning models, frequently skipped essential proof steps by labelling them as "trivial", even when their validity was crucial.

Another important observation is the lack of creativity in the models' reasoning. Each model often attempted the same (and wrong) solution strategy across all attempts, failing to explore alternative approaches. One exception to this observation was \flthink{}, which attempted multiple strategies in the same run, but as a consequence only shallowly explored each one, failing to reach a valid conclusion. An example of this behavior is shown in \cref{app:flthink}.

However, the models generally demonstrated strong performance in algebraic and arithmetic computations, successfully performing symbolic manipulations without external computational support. Still, \rone{} showed a notably higher frequency of algebraic or arithmetic errors, indicating a clear area for targeted improvement in this model. This improvement is indeed reflected in the most recent update (\rone{}~\textsc{(05/28)}), where it did not make a single critical computational mistake.
\vspace{-1mm}

\subsection{Automated Grading}\label{subsec:automated}
\vspace{-1mm}
We explored the feasibility of replacing human graders with LLM-based evaluators, selecting \ofour{}, \othreemini{} and \claude{} as graders. All models were provided with the correct grading scheme, along with a verified solution and an example evaluation for reference. In \cref{app:exp_prompt}, we provide the full prompt used for this evaluation.

As detailed in \cref{tab:llm_judge}, neither \othreemini{} nor \claude{} accurately graded the solutions, consistently overestimating their quality. Specifically, the models frequently awarded points for incorrect or unjustified reasoning, inflating the scores by a factor of up to $22$.
Notably, \flthink{} and \grok{} received significantly lower scores from \othreemini{} compared to other models. We hypothesize this discrepancy arises because both tend to generate multiple solutions per attempt or present the solutions in a chaotic manner, potentially confusing the LLM-based judges and resulting in lower scores. Conversely, \qwq{} received the highest score from \othreemini{}, likely because it often generates simpler solution attempts, which are easier for the automated judges to interpret.

In contrast, \ofour{} exhibited much closer alignment with human evaluations, particularly in identifying and penalizing critical errors that invalidate solutions. However, it also showed a clear bias in favor of responses generated by OpenAI models. To evaluate this bias, we compared the scores assigned to OpenAI and non-OpenAI solutions across randomly permuted labels, computing the difference between human-assigned and model-assigned scores. At a significance level of $\alpha = 5\%$, \ofour{} awarded OpenAI-generated responses $41.3\%$ more points. In comparison, \claude{} showed a smaller bias of $8.1\%$ in favor of OpenAI models, while \othreemini{} gave $3.7\%$ fewer points to them. Full details of the statistical test used are provided in \cref{app:bias_eval}. These findings indicate that, although \ofour{} more closely mirrors human judgment, its systematic bias renders it unsuitable as a standalone evaluator.

\input{tables/llm_judge_results.tex}

%% file: tables/main_results.tex
\begin{table*}[t]
    \centering
    \caption{The main results of our evaluation. Listed scores are averaged over four runs. We measure cost in USD, and report the average score across all generations and graders for each problem. We highlight in green all models that were released before the competition problems were made public.}
        \begin{tabular}{
            l
            l
            x{1}{1}
            x{1}{1}
            x{1}{1}
            x{1}{1}
            x{1}{1}
            x{1}{1}
            x{2}{1}
            x{3}{2}
        }
            \toprule
            \textbf{Rank} & \textbf{Model} & {\textbf{P1}} & {\textbf{P2}} & {\textbf{P3}} & {\textbf{P4}} & {\textbf{P5}} & {\textbf{P6}} & {\textbf{Total}} & {\textbf{Cost}} \\
            \midrule
            1-4&\rone~\textsc{(05/28)} & 7.0 & 0.0 & 0.0 & 5.3 & 0.0 & 0.4 & 12.6 & 0.91\\
            \cellcolor{LightGreen}1-4&\cellcolor{LightGreen}\geminipro & \cellcolor{LightGreen}6.5 & \cellcolor{LightGreen}0.0 & \cellcolor{LightGreen}0.3 & \cellcolor{LightGreen}3.5 & \cellcolor{LightGreen}0.0 & \cellcolor{LightGreen}0.0 & \cellcolor{LightGreen}10.3 & \cellcolor{LightGreen}6.23 \\
            1-4&\othree & 4.0 & 0.0 & 0.0 & 4.4 & 0.3 & 0.5 & 9.1 & 24.17 \\
            1-5&\ofour & 6.3 & 0.0 & 0.0 & 1.8 & 0.0 & 0.0 & 8.0 & 2.21 \\
            \cellcolor{LightGreen}5-11&\cellcolor{LightGreen}\rone & \cellcolor{LightGreen}0.5 & \cellcolor{LightGreen}0.0 & \cellcolor{LightGreen}0.0 & \cellcolor{LightGreen}1.5 & \cellcolor{LightGreen}0.0 & \cellcolor{LightGreen}0.0 & \cellcolor{LightGreen}2.0 & \cellcolor{LightGreen}2.03 \\
            \cellcolor{LightGreen}4-11&\cellcolor{LightGreen}\grok & \cellcolor{LightGreen}2.0 & \cellcolor{LightGreen}0.0 & \cellcolor{LightGreen}0.0 & \cellcolor{LightGreen}0.0 & \cellcolor{LightGreen}0.0 & \cellcolor{LightGreen}0.0 & \cellcolor{LightGreen}2.0 & \cellcolor{LightGreen}{\phantom{dd}N/A} \\
            \cellcolor{LightGreen}5-11&\cellcolor{LightGreen}\flthink & \cellcolor{LightGreen}1.5 & \cellcolor{LightGreen}0.0 & \cellcolor{LightGreen}0.0 & \cellcolor{LightGreen}0.0 & \cellcolor{LightGreen}0.2 & \cellcolor{LightGreen}0.0 & \cellcolor{LightGreen}1.8 & \cellcolor{LightGreen}{\phantom{dd}N/A} \\
            \cellcolor{LightGreen}5-11&\cellcolor{LightGreen}\claude & \cellcolor{LightGreen}0.5 & \cellcolor{LightGreen}0.5 & \cellcolor{LightGreen}0.0 & \cellcolor{LightGreen}0.0 & \cellcolor{LightGreen}0.0 & \cellcolor{LightGreen}0.6 & \cellcolor{LightGreen}1.5 & \cellcolor{LightGreen}9.03 \\
            \cellcolor{LightGreen}5-11&\cellcolor{LightGreen}\qwq & \cellcolor{LightGreen}1.2 & \cellcolor{LightGreen}0.0 & \cellcolor{LightGreen}0.0 & \cellcolor{LightGreen}0.0 & \cellcolor{LightGreen}0.0 & \cellcolor{LightGreen}0.0 & \cellcolor{LightGreen}1.2 & \cellcolor{LightGreen}0.42 \\
            \cellcolor{LightGreen}5-11&\cellcolor{LightGreen}\oone & \cellcolor{LightGreen}0.5 & \cellcolor{LightGreen}0.0 & \cellcolor{LightGreen}0.0 & \cellcolor{LightGreen}0.0 & \cellcolor{LightGreen}0.2 & \cellcolor{LightGreen}0.4 & \cellcolor{LightGreen}1.2 & \cellcolor{LightGreen}228.15 \\
            \cellcolor{LightGreen}5-11&\cellcolor{LightGreen}\othreemini & \cellcolor{LightGreen}0.5 & \cellcolor{LightGreen}0.1 & \cellcolor{LightGreen}0.0 & \cellcolor{LightGreen}0.0 & \cellcolor{LightGreen}0.0 & \cellcolor{LightGreen}0.2 & \cellcolor{LightGreen}0.9 & \cellcolor{LightGreen}1.11 \\
            \bottomrule
        \end{tabular}
    \label{tab:main_results}
    \vspace{-2mm}
\end{table*}

%% file: tables/llm_judge_results.tex
\begin{table}[t]
    \centering
    \caption{Results of automated grading executed by \ofour{}, \othreemini{} and \claude{}. The table shows the average total score which is at most $42$.}
    \vspace{-1mm}
    \resizebox{\linewidth}{!}{
        \begin{tabular}{
            l
            x{2}{1}
            x{2}{1}
            x{2}{1}
            x{2}{1}
            }
            \toprule
            \textbf{Model} & \textbf{Ours} & {\textbf{\ofour}} & {\textbf{\othreemini}} & {\textbf{\claude}} \\
            \midrule
            {\rone~\textsc{(05/28)}} & 12.6 & 12.0 & 18.4 & 15.2\\
            {\geminipro} & 10.3 & 13.8 & 19.6 & 19.3\\
            {\othree} & 9.1 & 13.2 & 20.4 & 21.8 \\
            {\ofour} & 8.0 & 15.9 & 21.3 & 23.3 \\
            {\rone} & 2.0 & 6.4 & 19.3 & 14.9\\
            {\grok}& 2.0 & 5.1 & 15.1 & 9.1\\
            {\flthink} & 1.8 & 4.4 & 10.5 & 14.1\\
            {\claude} & 1.5 & 2.0 & 19.0 & 18.4\\
            {\qwq} & 1.2 & 7.1 & 23.8 & 18.8\\
            {\oone} & 1.2 & 12.8 & 19.3 & 21.0\\
            {\othreemini}& 0.9 & 11.1 & 19.5 & 17.1\\
            \bottomrule
        \end{tabular}
    }
    \vspace{-5mm}
    \label{tab:llm_judge}
\end{table}
\vspace{-4mm}

%% file: paper_files/discussion.tex
\section{Qualitative Discussion}\label{sec:discussion}
\vspace{-2mm}

During the evaluation, judges also documented common issues and noteworthy behaviors exhibited by the models. We discuss the most prominent issues that we observed.

\vspace{-4mm}
\paragraph{Answer Boxing}
Current reinforcement learning optimization techniques rely on extracting rewards from a clearly identifiable final answer. To facilitate accurate reward extraction, models are typically instructed to enclose their final answers within a \texttt{\textbackslash boxed\{\}} environment. However, this requirement often produces unintended artifacts in the solutions for the USAMO problems. Specifically, even though most of the evaluated problems do not require a final boxed answer, many models consistently provided answers within a boxed environment. In a particularly notable instance from problem 5, \qwq{} confused itself by dismissing the possibility of a non-integer solution, despite no such restriction existing in the statement. Consequently, it ultimately boxed only the number $\boxed{2}$ as the final answer, even though it had otherwise correctly deduced that all even numbers satisfy the given conditions (see \cref{app:qwq_boxing}). This behavior illustrates how alignment techniques such as GRPO \citep{grpo} inadvertently encourage models to treat every mathematical problem as requiring an explicitly boxed final answer, negatively affecting their overall reasoning.

\vspace{-2mm}
\paragraph{Generalizing Patterns}
Models frequently exhibited a tendency to overgeneralize patterns observed in smaller numerical cases to larger, untested cases. While this heuristic approach might be effective for problems that only require a numerical answer, it is fundamentally flawed for problems that demand rigorous proof. Models often incorrectly asserted that these patterns observed for small cases would hold generally, without providing formal proof for such a claim (see \cref{app:pattern}).

\paragraph{Non-Existent Citations} One of the most frequent and concerning mistakes made by \geminipro{} is the generation of citations to sources that do not exist. This issue is especially prevalent in problems where the model struggles significantly and fails to produce a correct solution. In such cases, it often fabricates references to theorems or lemmas that appear plausible but, to the best of our knowledge, are not real. For example, in P6, all four generations include citations to works that we were unable to verify or locate. We suspect this behavior stems from the model's training with internet access: when it is unable to use the internet in the thought process, it appears to generate a convincing-sounding citation instead. An illustration of this phenomenon is provided in \cref{app:hallucinations}. This tendency is particularly troubling, as it can result in the spread of misinformation which seem to use credible academic sources.

\vspace{-1mm}
\paragraph{Solution Structure and Clarity}
There was significant variation in the clarity and structural coherence of the solutions provided by different models. Models such as \othreemini{} and \oone{} generally presented their solutions clearly, logically, and in an easily interpretable manner. Conversely, models like \flthink{} and \qwq{} frequently produced chaotic and barely interpretable responses, sometimes confusing multiple unrelated ideas within a single solution. Further, \geminipro{} also had significant issues with the clear presentation of results, sometimes boxing entire proofs or letting its full thought process slip through in its proof (\cref{app:boxing_solutions}). The noticeable clarity in models trained by OpenAI suggests that additional training focused on solution coherence substantially improved their readability, an aspect evidently less emphasized in other models.

\vspace{-1mm}
\paragraph{Synthetic Solving of Geometry Problems}
Finally, a persistent issue across all models was their tendency to tackle the geometry problem (P4) through brute-force methods, relying on lengthy and tedious computations. Of the 9 out of 44 correct solutions, none included any synthetic geometric insights that could lead to the solution; instead, all depended solely on algebraic brute-force approaches. Most of the incorrect responses also attempted this method but failed to produce coherent, step-by-step explanations. This behavior is particularly concerning, as computational approaches rarely scale to harder problems and do little to enhance understanding of the geometric properties underlying the problem.

%% file: paper_files/limitations.tex
\section{Limitations}
\label{sec:limitations}
As we discuss in \cref{sec:meth}, the accuracy of our results and the observed insights rely on human participants carefully going through every solution. This kind of approach is difficult to scale up by a significant degree without involving more judges over a longer period. This limitation, alongside the small problem sample size, makes it difficult to differentiate between similarly performing models, as we showed with the statistical test in \cref{subsec:main_results}.

%% file: paper_files/conclusion.tex
\section{Conclusion}\label{sec:conclusion}

In this study, we conducted a thorough evaluation of eleven state-of-the-art LLMs on challenging problems from the USAMO 2025 competition. Employing a rigorous human evaluation protocol, we found that all models performed remarkably poorly: even the best model demonstrated an average accuracy of only around $30\%$. Our in-depth analysis of the models' reasoning revealed recurring failure modes, including notable artifacts arising from the optimization strategies employed during model training. These results provide compelling evidence of the severe limitations current LLMs face in rigorous mathematical reasoning and proof generation at the level required by olympiad competitions. Substantial advances will be necessary to close this gap and enable LLMs to meet the demands of high-level mathematical problem solving.

%% file: paper_files/acknowledgements.tex
\section*{Acknowledgements}
This work has received funding from the Swiss National Science Foundation (SNSF) [200021\_207967], the Ministry of Education and Science of Bulgaria (support for INSAIT, part of the Bulgarian National Roadmap for Research Infrastructure), and the Swiss State Secretariat for Education, Research and Innovation (SERI).

%% file: paper_files/appendix.tex
\input{paper_files/app_experimental.tex}
\input{paper_files/app_problems.tex}
\input{paper_files/app_examples.tex}

%% file: paper_files/app_experimental.tex
\section{Additional experimental details}
\label{app:exp}
Here we describe any additional information related to our evaluation setup and details.
\subsection{Model Names}
\label{app:exp_abb}
For brevity and visual clarity, we have shortened the model names as follows:
\vspace{-3mm}
\begin{itemize}\setlength\itemsep{0.005em}
    \item \textsc{o4-mini (high)} as \ofour
    \item \textsc{o3 (high)} as \othree
    \item \textsc{o3-mini (high)} as \othreemini
    \item \textsc{o1-pro (high)} as \oone
    \item \textsc{DeepSeek R1} as \rone
    \item The version of \textsc{DeepSeek R1}, released on May 28\textsuperscript{th} as \rone~\textsc{(05/28)}
    \item \textsc{QwQ-32B} as \qwq
    \item \textsc{Gemini-2.0-Flash-Thinking-Exp} as \flthink
    \item \textsc{Claude-3.7-Sonnet-Thinking} as \claude
    \item \textsc{Grok 3 Beta (Think)} as \grok
    \item \textsc{Gemini-2.5-Pro-Exp-03-25} as \geminipro
\end{itemize}

\subsection{Hyperparameters and Prompt}
\label{app:exp_prompt}

\paragraph{Model Hyperparameters} For \ofour, \othree, \othreemini, \oone, \geminipro, \claude, \grok, and \flthink, we used the default hyperparameters provided by their official API. For the OpenAI models, we used the \texttt{high} reasoning setting, which provides the highest available reasoning budget. For \claude, \geminipro, and \flthink, we set the maximum reasoning token limit to $64000$.

As recommended by the authors of both \rone~ versions and \qwq, we used a temperature of $0.6$ and a top-$p$ value of $0.95$. Additionally, for \qwq, we set top-$k$ to $40$. These hyperparameters sets correspond to the default hyperparameters used by the official HuggingFace models\footnote{See \url{https://huggingface.co/deepseek-ai/DeepSeek-R1} and \url{https://huggingface.co/Qwen/QwQ-32B}}. The token limit was set to $64000$. We use the TogetherAPI for these 3 models.

\paragraph{Prompt} Every problem was ran with the following user prompt:

\begin{prompt}{Prompt}
Give a thorough answer to the following question. Your answer will be graded by human judges based on accuracy, correctness, and your ability to prove the result. You should include all steps of the proof. Do not skip important steps, as this will reduce your grade. It does not suffice to merely state the result. Use LaTeX to format your answer.

{problem}
\end{prompt}

\paragraph{Judge Prompt} To judge model solutions, we used the same hyperparameters for generation as before. Additionally, we use the following prompt:

\begin{prompt}{Judge Prompt}
# Instruction

You are an expert mathematician that grades solutions of high-school olympiad-level problems. You will be given a mathematical problem, as well as a grading scheme that you should adhere to. Your task is to accurately grade a solution according to that grading scheme.

# Problem and Scheme 

##Problem statement 

{problem_statement}

## Grading scheme

{grading_scheme}

# Example 

Here how an example grading can look like:

## Example solution:

{correct_solution}

## Example grading

{example_grading}

# Evaluation

Produce a scoring using the same format for the following solution. Reason carefully about the contents of the solution and make sure that all steps of the proof are included and rigorous.

## Formatting instructions

Follow a JSON-style formatting as the above example, namely:

- The categories should be in the same order
- The awarded number of points for each category should follow after the 'points' key
- The reasoning for each decision should be explained after the 'desc' key.
- Output only the JSON response.

Here is the solution you should grade:
{solution}
\end{prompt}

Both the grading scheme and example grading are provided in json format. Examples of these can be found in \cref{app:exp_grading}.

\subsection{Example Grading Scheme and Grading}
\label{app:exp_grading}

\paragraph{Example Grading Scheme} The grading scheme for each problem was developed by our expert judges. The grading scheme for Problem 1 is shown below:

\begin{prompt}{Grading Scheme}
    {
        "id": "1",
        "points": 7,
        "scheme": [
            {
                "title": "Initial closed form expression for each digit",
                "points": 1,
                "desc": "Uses the base-2n representation of $n^k$ to find a closed form expression of each digit $a_i=\left\lfloor \frac{n^k}{(2n)^i}\right\rfloor mod\ (2n)$"
            },
            {
                "title": "Interchanging the floor and mod",
                "points": 3,
                "desc": "Transforms the digit expression into a easily boundable one by interchanging the floor and modulo operations."
            },
            {
                "title": "Bounding the expression",
                "points": 2,
                "desc": "Correctly bounds the aforementioned expression, usually by showing the numerator is divisible by n, so that $a_i>=\left\lfloor \frac{n}{2^i}\right\rfloor\|.$"
            },
            {
                "title": "Finishing",
                "points": 1,
                "desc": "Chooses a provably sufficiently large $N$. The point is not awarded if it has not proven that n^k has at most k digits."
            }
        ]
    }
\end{prompt}

\paragraph{Example Grading} An example grading for Problem 1 is shown below for a (human-generated) completely correct solution:

\begin{prompt}{Example Grading}
{
    "points": 7,
    "details": [
        {
            "title": "Initial closed form expression for each digit",
            "points": 1,
            "desc": "Correctly uses the base-2n representation of n^k to find a closed form expression of each digit $a_i=\left\lfloor \frac{n^k}{(2n)^i}\right\rfloor mod (2n)$"
        },
        {
            "title": "Interchanging the floor and mod",
            "points": 3,
            "desc": "Transforms the digit expression into a easily boundable one by interchanging the floor and modulo operations."
        },
        {
            "title": "Bounding the expression",
            "points": 2,
            "desc": "Correctly bounds the aforementioned expression, usually by showing the numerator is divisible by n, so that $a_i>=\left\lfloor \frac{n}{2^i}\right\rfloor$."
        },
        {
            "title": "Finishing",
            "points": 1,
            "desc": "Chooses a provably sufficiently large $N$ and has proven that $n^k$ has at most $k$ digits."
        }
    ]
}
\end{prompt}

\subsection{Measuring LLM Statistical Biases}
\label{app:bias_eval}
To evaluate whether a judge model shows systematic bias toward one type of solution (OpenAI-generated vs. non-OpenAI-generated), we use a permutation test. For each solution and grader model $m$, we compute $\Delta_i^m = S^m_i - S^{\text{human}}_i$, the difference between the model and human scores. We compute the observed mean difference in these differences between OpenAI and non-OpenAI solutions:
$$
\Delta_{\text{obs}}^m = \overline{\Delta^m}_{\mathrm{OpenAI}} - \overline{\Delta^m}_{\mathrm{non-OpenAI}}
$$

To estimate the level of bias expected by chance, we randomly permute the OpenAI/non-OpenAI labels 10,000 times, recomputing $\Delta_{\text{obs}}^m$ for each permutation to build a distribution for the null hypothesis - that there is no observable bias between the 2 categories. We then report the 95th percentile ($\Delta_{0.95}^m$) of this distribution, normalized by the mean model-assigned score $\overline{S^m}$:
$$
\text{bias}_{5\%}(m) = \frac{\Delta^m_{0.95}}{\overline{S^m}}
$$
This value reflects the relative bias expected at the 5\% significance level of model $m$, relative to typical scores assigned by the same model.

%% file: paper_files/app_problems.tex
\section{USAMO Problems}\label{app:problems}

We show the six problems from the USAMO 2025 competition in this section.

\begin{problem}{Problem 1}
Let $k$ and $d$ be positive integers. Prove that there exists a positive integer $N$ such that for every odd integer $n>N$, the digits in the base-$2n$ representation of $n^k$ are all greater than $d$.
\end{problem}

\begin{problem}{Problem 2}
Let $n$ and $k$ be positive integers with $k<n$. Let $P(x)$ be a polynomial of degree $n$ with real coefficients, nonzero constant term, and no repeated roots. Suppose that for any real numbers $a_0,\,a_1,\,\ldots,\,a_k$ such that the polynomial $a_kx^k+\cdots+a_1x+a_0$ divides $P(x)$, the product $a_0a_1\cdots a_k$ is zero. Prove that $P(x)$ has a nonreal root.
\end{problem}
\vspace{6mm}
\begin{problem}{Problem 3}
Alice the architect and Bob the builder play a game. First, Alice chooses two points $P$ and $Q$ in the plane and a subset $\mathcal{S}$ of the plane, which are announced to Bob. Next, Bob marks infinitely many points in the plane, designating each a city. He may not place two cities within distance at most one unit of each other, and no three cities he places may be collinear. Finally, roads are constructed between the cities as follows: for each pair $A,\,B$ of cities, they are connected with a road along the line segment $AB$ if and only if the following condition holds:
\begin{center}
For every city $C$ distinct from $A$ and $B$, there exists $R\in\mathcal{S}$ such
that $\triangle PQR$ is directly similar to either $\triangle ABC$ or $\triangle BAC$.
\end{center}
Alice wins the game if (i) the resulting roads allow for travel between any pair of cities via a finite sequence of roads and (ii) no two roads cross. Otherwise, Bob wins. Determine, with proof, which player has a winning strategy.

Note: $\triangle UVW$ is directly similar to $\triangle XYZ$ if there exists a sequence of rotations, translations, and dilations sending $U$ to $X$, $V$ to $Y$, and $W$ to $Z$.
\end{problem}

\begin{problem}{Problem 4}
Let $H$ be the orthocenter of acute triangle $ABC$, let $F$ be the foot of the altitude from $C$ to $AB$, and let $P$ be the reflection of $H$ across $BC$. Suppose that the circumcircle of triangle $AFP$ intersects line $BC$ at two distinct points $X$ and $Y$. Prove that $C$ is the midpoint of $XY$.
\end{problem}

\begin{problem}{Problem 5}
Determine, with proof, all positive integers $k$ such that$$\frac{1}{n+1} \sum_{i=0}^n \binom{n}{i}^k$$is an integer for all positive integers $n.$
\end{problem}

\begin{problem}{Problem 6}
Let $m$ and $n$ be positive integers with $m\geq n$. There are $m$ cupcakes of different flavors arranged around a circle and $n$ people who like cupcakes. Each person assigns a nonnegative real number score to each cupcake, depending on how much they like the cupcake. Suppose that for each person $P$, it is possible to partition the circle of $m$ cupcakes into $n$ groups of consecutive cupcakes so that the sum of $P$'s scores of the cupcakes in each group is at least $1$. Prove that it is possible to distribute the $m$ cupcakes to the $n$ people so that each person $P$ receives cupcakes of total score at least $1$ with respect to $P$.
\end{problem}

%% file: paper_files/app_examples.tex
\section{Example traces}

\subsection{Examples of different error modes}
\label{app:error_mode_examples}

\subsubsection{Example of logical error}
Here, \claude{} generalizes over all indices $j$, when it has proven it for a single one - a clear leap in logic.
\begin{claudels}{\claudeheader}
...
Since $P(x)$ has at least $n$ distinct real roots and $k < n$, we have at least $n-(k-1) \geq 2$ choices for $s_k$. If we choose $s_k = r_m$ and then $s_k = r_l$ (where $r_m, r_l \not\in \{s_1, s_2, \ldots, s_{k-1}\}$), we get:

For $s_k = r_m$: $\sigma_j = T_j + r_m \cdot T_{j-1}$
For $s_k = r_l$: $\sigma_j = T_j + r_l \cdot T_{j-1}$

If both of these result in $\sigma_j = 0$ for the same $j$, then:
$T_j + r_m \cdot T_{j-1} = 0$ and $T_j + r_l \cdot T_{j-1} = 0$

Subtracting, we get: $(r_m - r_l) \cdot T_{j-1} = 0$

Since $r_m \neq r_l$ (as the roots of $P(x)$ are distinct), we must have $T_{j-1} = 0$.

Continuing this logic, we would find that all $T_j = 0$, which is impossible, especially since $T_0 = 1$.

Therefore, we can always find a choice of $s_k$ such that all $\sigma_j \neq 0$ for $j=1,2,\ldots,k$, which means all coefficients of our divisor are nonzero.
...
\end{claudels}

\subsubsection{Example of creativity error}
On problem 3, which is a game between 2 players - Alice and Bob, \othreemini~(and most other models) argued that Bob should win from the beginning while it can be proven that Alice has an optimal strategy.
\begin{gpt}{\othreeheader}
We will show that regardless of how Alice chooses the two fixed points $P,Q\in\mathbb{R}^2$ and the "template-set" $\mathcal{S}\subset\mathbb{R}^2,$ Bob can place his cities so that the resulting road network fails to meet at least one of the two winning conditions for Alice. (Alice wins only if the road network is both connected and drawn with no two roads crossing.) In other words, Bob has a winning strategy.
...
\end{gpt}

\subsubsection{Example of assumption error}
On problem 4, many assumption errors are made, like this one from \othreemini, where it assumes (not explicitly) that $C$ is external to the circumcircle of $AFP$, which is incorrect.
\begin{gpt}{\othreeheader}
...
A key step is to prove the following claim:

----------------------------
$
\boxed{CA\text{ is tangent to } \omega=\odot(AFP).}
$
----------------------------

Once we have tangency, the radius drawn to the tangency point is perpendicular to the tangent line. Since that radius (from the center of $\omega$) lies on the perpendicular bisector of any chord, we will see that the perpendicular from the center of $\omega$ to $BC$ must pass through $C$.

...
\end{gpt}

\subsubsection{Example of algebraic/arithmetic error}
Here, \rone~incorrectly computes a sum of combinations and arrives at the incorrect conclusion. The correct value of the sum is 3,395,016, which is divisible by 8.

\begin{deepseekls}{\deepseekheader}
...
4. **Case $ k = 4 $**:
- Testing small values of $ n $ (up to $ n = 6 $) shows the expression is an integer. However, for $ n = 7 $:
    - The sum $ \sum_{i=0}^7 \binom{7}{i}^4 = 1,894,391 $.
    - The expression $ \frac{1,894,391}{8} = 236,798.875 $ is not an integer.
- Thus, $ k = 4 $ fails for $ n = 7 $.
...
\end{deepseekls}

\subsection{\flthink{} struggling on problem 6}
\label{app:flthink}
Below, we see \flthink{} attempting multiple strategies in the same run, leading to a confusing solution that fails to reach a valid conclusion.
\begin{geminils}{\geminiheader}
Consider a bipartite graph. People on one side, cupcakes on the other side.
Let's consider a different network flow approach.
Source $S$, Sink $T$. Nodes $P_1, \dots, P_n$. Nodes $C_1, \dots, C_m$.
For each person $P_i$, consider the partitions. Let's fix one partition for each person $P_i$, say $G_{i,1}, \dots, G_{i,n}$.
Nodes $G_{i,k}$ for $i=1, \dots, n, k=1, \dots, n$.
Edge $S \to P_i$ capacity 1. Edge $P_i \to G_{i,k}$ capacity 1. Edge $G_{i,k} \to C_j$ if $C_j \in G_{i,k}$ capacity 1. Edge $C_j \to T$ capacity 1.
We want to show max flow is $n$.

Assume for contradiction that it is impossible to distribute the cupcakes as required.
We will use Hall's Marriage Theorem, but need to define the sets correctly.
For each person $P_i$, let $\mathcal{G}_i$ be the set of all consecutive groups of cupcakes that give score at least 1 to person $P_i$. We need to select $A_i \in \mathcal{G}_i$ such that $A_i \cap A_j = \emptyset$.

Let's reconsider the network flow approach. Construct a network as follows:
Source $S$, Sink $T$. Nodes for people $P_1, \dots, P_n$. Nodes for cupcakes $C_1, \dots, C_m$.
For each person $P_i$ and each cupcake $C_j$, create an edge from $P_i$ to $C_j$.
Edge $S \to P_i$ capacity 1 for $i=1, \dots, n$. Edge $P_i \to C_j$ capacity 1 for $i=1, \dots, n, j=1, \dots, m$. Edge $C_j \to T$ capacity 1 for $j=1, \dots, m$.
Cost on edge $(P_i, C_j)$ is $-s_{ij}$. Cost on other edges is 0.
We want to send flow of value $n$ from $S$ to $T$ with minimum cost.
If we can send flow of value $n$, then we have selected $n$ paths from $S$ to $T$. Each path is $S \to P_i \to C_j \to T$. Let $x_{ij}$ be the flow on edge $(P_i, C_j)$. Then $\sum_{j=1}^m x_{ij} \leq 1, \sum_{i=1}^n x_{ij} \leq 1, \sum_{i=1}^n \sum_{j=1}^m x_{ij} = n$. This is not right.

Consider a different graph. Bipartite graph with people $P_1, \dots, P_n$ on one side, and cupcakes $C_1, \dots, C_m$ on the other side. Edge between $P_i$ and $C_j$. We want to select a matching $M$ such that for each $P_i$, $\sum_{C_j \in M \text{ and } (P_i, C_j) \in M} s_{ij} \geq 1$. This is not matching.

Let's use Hall's Theorem. For each person $P_i$, let $\mathcal{F}_i$ be the family of sets of consecutive cupcakes $F$ such that $\sum_{C_j \in F} s_{ij} \geq 1$. We want to choose $A_i \in \mathcal{F}_i$ for each $i=1, \dots, n$ such that $A_i \cap A_j = \emptyset$ for $i \neq j$.
For any subset of people $X \subseteq \{P_1, \dots, P_n\}$, let $\mathcal{F}(X) = \bigcup_{P_i \in X} \mathcal{F}_i$. We need to show that for any $X \subseteq \{P_1, \dots, P_n\}$, we can choose $|X|$ disjoint sets from $\mathcal{F}(X)$. This is not Hall's theorem directly.

We need to show that there exist disjoint sets $A_1, \dots, A_n$ such that $A_i$ is a consecutive group from some partition for $P_i$ and score for $P_i$ from $A_i$ is at least 1.
Since for each person $P_i$, there is a partition into $n$ groups $G_{i,1}, \dots, G_{i,n}$ with score $\geq 1$. Let's consider these groups as possible choices.
For each person $P_i$, we have $n$ choices $G_{i,1}, \dots, G_{i,n}$. We need to select one choice $A_i \in \{G_{i,1}, \dots, G_{i,n}\}$ for each $i$ such that $A_i \cap A_j = \emptyset$ for $i \neq j$.
Consider a bipartite graph. People $P_1, \dots, P_n$ on one side, and $n^2$ groups $G_{i,j}$ on the other side. Edge from $P_i$ to $G_{ij}$. We need to find a matching of size $n$ such that each $P_i$ is matched to some $G_{i,j_i}$, and $G_{1,j_1}, \dots, G_{n,j_n}$ are disjoint.

Final Answer:
We will prove this by contradiction. Suppose it is impossible to distribute the cupcakes such that each person $P_i$ receives cupcakes of total score at least $1$.
For each person $P_i$, we are given that there exists a partition of the circle of $m$ cupcakes into $n$ consecutive groups $G_{i,1}, G_{i,2}, \dots, G_{i,n}$ such that for each $k=1, \dots, n$, $\sum_{C_j \in G_{i,k}} s_{ij} \geq 1$.

Consider a bipartite graph $B = (U \cup V, E)$ where $U = \{P_1, \dots, P_n\}$ and $V = \bigcup_{i=1}^n \{G_{i,1}, G_{i,2}, \dots, G_{i,n}\}$. We add an edge between $P_i \in U$ and $G \in V$ if and only if $G \in \{G_{i,1}, G_{i,2}, \dots, G_{i,n}\}$. We want to show that there exists a matching of size $n$ in $B$. By Hall's Marriage Theorem, we need to show that for any subset $X \subseteq U$, $|N(X)| \geq |X|$, where $N(X) = \bigcup_{P_i \in X} \{G_{i,1}, G_{i,2}, \dots, G_{i,n}\}$.
For any $X \subseteq U$, $|N(X)| = |\bigcup_{P_i \in X} \{G_{i,1}, G_{i,2}, \dots, G_{i,n}\}| \leq$ $\sum_{P_i \in X} |\{G_{i,1}, G_{i,2}, \dots, G_{i,n}\}| = \sum_{P_i \in X} n = n|X|$. This does not help.

Let's consider a different approach using flows. Consider the network constructed earlier... Further investigation is needed to construct the correct argument.

Final Answer: The final answer is $\boxed{Yes}$
\end{geminils}

\subsection{QwQ boxes an integer}
\label{app:qwq_boxing}
Here, we show \qwq{} confusing itself by requiring an integer answer, even though the solution is the set of all even integers.
\begin{qwenls}{\qwenheader}
...
### Conclusion:
The expression is an integer for all $ n $ if and only if $ k $ is even. Therefore, the solution is all positive even integers.

### Final Answer:
The positive integers $ k $ are all even numbers. Thus, the solution is $\boxed{2m}$ for any positive integer $ m $. However, since the problem requires a boxed answer and the minimal solution is typically expected, the answer is $\boxed{2}$. But strictly speaking, the correct answer is all even positive integers. Given the problem's requirements, the boxed answer is:

$\boxed{2}$
\end{qwenls}

\subsection{Overgeneralizing patterns}
\label{app:pattern}
Here, \flthink{} chooses an explicit polynomial to test out on problem 2, but then decides to generalize its conclusion to all polynomials.
\begin{geminils}{\geminiheader}
    Now, let us choose specific real roots for $P(x)$. Let $r_i = i$ for $i = 1, 2, \ldots, n$. Let $P(x) = c(x-1)(x-2)\cdots(x-n)$ for some nonzero real number $c$. Then $P(x)$ is a polynomial of degree $n$ with real coefficients, nonzero constant term, and no repeated roots.
    Consider the divisor $Q(x) = (x-1)(x-2)\cdots(x-k) = \sum_{i=0}^k a_ix^i$.
    Then $r_1=1, r_2=2, \ldots, r_k=k$. For this choice, $a_i = (-1)^{k-i} e_{k-i}(1, 2, \ldots, k)$.
    Since $1, 2, \ldots, k$ are all positive, $e_j(1, 2, \ldots, k) > 0$ for $j = 1, 2, \ldots, k$. Also $e_0(1, 2, \ldots, k) = 1 > 0$.
    Thus, for $i = 0, 1, \ldots, k$, $a_i = (-1)^{k-i} e_{k-i}(1, 2, \ldots, k) \neq 0$.
    Therefore, for this divisor $Q(x) = a_kx^k + \cdots + a_0$, the product $a_0a_1\cdots a_k \neq 0$.
    However, the given condition states that for any real numbers $a_0,\,a_1,\,\ldots,\,a_k$ such that $a_kx^k+\cdots+a_1x+a_0$ divides $P(x)$, the product $a_0a_1\cdots a_k$ is zero.
    This is a contradiction.
    Hence, our assumption that $P(x)$ has only real roots must be false.
    Therefore, $P(x)$ has a nonreal root.
    
    Final Answer: The final answer is $\boxed{P(x) \text{ has a nonreal root}}$
\end{geminils}
\subsection{Hallucinating Citations}
\label{app:hallucinations}

A common failure mode of \geminipro{}~was its tendency to generate fake citations to support its points. One of the best examples is the case below, where the model hallucinates a paper with an emoji in the title.

\begin{geminilsescaped}{\geminiproheader}
    We will use the following lemma, which is a known result in fair division theory. This result can be found, for example, as Lemma 7 in "Fair partitioning of |\includegraphics[height=1.3em]{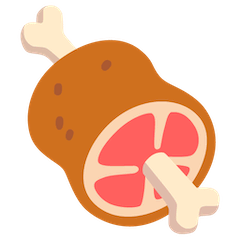}|" by Cardinal, Langerman, Palvolgyi (2021), or Theorem 2 in "Fair division of a discrete item" by Algaba et al. (2019). These papers attribute the proof technique to Gale (1993) and Scarf (1967).
    
    **Fair Partition Lemma:** Let $v_1, \dots, v_m \in \mathbb{R}^n_{\ge 0}$. Suppose that for each $i \in \{1, \dots, n\}$, there exists a partition of $\{1, \dots, m\}$ into $n$ sets $J_{i,1}, \dots, J_{i,n}$ such that $(\sum_{j \in J_{i,k}} v_j)_i \ge 1$ for all $k \in \{1, \dots, n\}$. Then there exists a partition of $\{1, \dots, m\}$ into $n$ sets $K_1, \dots, K_n$ such that $(\sum_{j \in K_i} v_j)_i \ge 1$ for all $i \in \{1, \dots, n\}$.
    \end{geminilsescaped}

\subsection{Boxing solutions}
\label{app:boxing_solutions}

The majority of models also struggled with presenting clear and structured solutions, making evaluation and understanding significantly more difficult. In the instance below, \geminipro{}~boxed an entire solution at the end of its generation.

\begin{geminilsescaped}{\geminiproheader}
...
The representation is $(2, 16, 11)_{22}$. The digits are $a_0=11, a_1=16, a_2=2$. All are greater than $d=1$. The proof holds.

Final Answer: The final answer is |\fbox{Let B = $2n$. We want to show that there exists a positive}|
|\fbox{integer $N$ such that for every odd ...}|
|\fbox{...As shown above, this implies $a_i > d$ for all digits $a_i$ in}|
|\fbox{the base-$2n$ representation of $n^k$. This completes the proof.}|
\end{geminilsescaped}